\documentclass[runningheads]{llncs}
\usepackage{graphicx}
\usepackage{comment}
\usepackage{soul}
\usepackage{amsmath,amssymb} % define this before the line numbering.
\usepackage{color}
\usepackage{amsmath}
\usepackage{kantlipsum}
\usepackage{float}
\usepackage{subfigure}

\usepackage{algorithm}
\usepackage{algorithmic}

\usepackage[width=122mm,left=12mm,paperwidth=146mm,height=193mm,top=12mm,paperheight=217mm]{geometry}

    \makeatletter
\def\@fnsymbol#1{\ensuremath{\ifcase#1\or \dagger\or \ddagger\or
   \mathsection\or \mathparagraph\or \|\or **\or \dagger\dagger
   \or \ddagger\ddagger \else\@ctrerr\fi}}
    \makeatother

\begin{document}
% \renewcommand\thelinenumber{\color[rgb]{0.2,0.5,0.8}\normalfont\sffamily\scriptsize\arabic{linenumber}\color[rgb]{0,0,0}}
% \renewcommand\makeLineNumber {\hss\thelinenumber\ \hspace{6mm} \rlap{\hskip\textwidth\ \hspace{6.5mm}\thelinenumber}}
% \linenumbers
\setlength{\abovedisplayskip}{4pt}
\setlength{\belowdisplayskip}{4pt}
\pagestyle{headings}
\mainmatter

\title{A new geodesic-based feature for characterization of 3D shapes: application to soft tissue organ temporal deformations}  

%\title{A novel geodesic-based feature for characterization of smooth organ dynamics} % Replace with your title

% INITIAL SUBMISSION 
%\begin{comment}
%\titlerunning{} 
\authorrunning{Makki et al.} 
\author{Karim Makki$^{1,*}$, Amine Bohi$^1$, Augustin C. Ogier$^1$, Marc Emmanuel Bellemare$^1$}
\institute{$^1$ Aix Marseille Univ, Universit\'{e} de Toulon, CNRS, LIS, Marseille, France}
%\end{comment}
%******************

% CAMERA READY SUBMISSION
\begin{comment}
\titlerunning{}
% If the paper title is too long for the running head, you can set
% an abbreviated paper title here
%
\author{Karim Makki \inst{1} \orcidID{0000-0003-1915-2488} ,
Amine Bohi\inst{1}\orcidID{0000-0002-2435-3017} ,
Augustin C. Ogier\inst{1}\orcidID{0000-0001-9178-9964} ,
Marc Emmanuel Bellemare\inst{1}\orcidID{0000-0001-9913-3097}}

%
\authorrunning{K. Makki et al.}
% First names are abbreviated in the running head.
% If there are more than two authors, 'et al.' is used.
%
\institute{Aix Marseille Univ, Universit\'{e} de Toulon, CNRS, LIS, Marseille, France
\email{karim.makki@univ-amu.fr}\\}
\end{comment}
%******************
\maketitle

\begin{abstract}

In this paper, we propose a method for characterizing 3D shapes from point clouds and we show a direct application on a study of organ temporal deformations. As an example, we characterize the behavior of a bladder during a forced respiratory motion with a reduced number of 3D surface points: first, a set of equidistant points representing the vertices of quadrilateral mesh for the surface in the first time frame are tracked throughout a long dynamic MRI sequence using a Large Deformation Diffeomorphic Metric Mapping (LDDMM) framework. Second, a novel geometric feature which is invariant to scaling and rotation is proposed for characterizing the temporal organ deformations by employing an Eulerian Partial Differential Equations (PDEs) methodology. We demonstrate the robustness of our feature on both synthetic 3D shapes and realistic dynamic MRI data portraying the bladder deformation during forced respiratory motions. Promising results are obtained, showing that the proposed feature may be useful for several computer vision applications such as medical imaging, aerodynamics and robotics. 


\end{abstract}

\section{Introduction}
Pelvic floor disorders affect approximately 50\% of women older than 50 years \cite{law2008mri}. Health problems related to these disorders such as urinary and fecal incontinences get worse with age which affects activities of daily living. To help surgeons and clinicians for better understanding these disorders, it is crucial to establish the normative pelvic floor dynamics in healthy women before focusing pathomechanics studies. In the literature, only a few studies have attempted to track pelvic organ deformations \textit{in vivo}. An ex-vivo study to evaluate deformations of the urinary bladder wall during whole bladder filling is presented in~\cite{parekh2010ex}. In~\cite{rios2017population}, the authors proposed a model to predict bladder motion and deformation between fractions. However, this study was based on computed tomographic (CT) scans which expose the patient to radiation (X-rays). This suggests that this imaging technique is not suitable for large-scale studies. In this paper, we propose to quantify and characterize the bladder deformations during a forced breathing exercise, using dynamic MRI. Dynamic MRI is a non-invasive imaging technique that has made it possible to explore the pelvic floor system during respiratory motion. However, dynamic MRI suffers from its low-resolution in addition to its sensitivity to motion artifacts. A diffeomorphic mapping based characterization of temporal sequences is first presented in~\cite{rahim2013diffeomorphic} to quantify the temporal deformation of the pelvic floor organs in 2D+t. However, a more informative 3D+t quantification of the pelvic floor organ dynamics is necessary because of the large shape variability of the pelvic organs across time. To overcome these limitations, a combination of spatial resolution of conventional static MRI and temporal resolution of dynamic MRI data is necessary. First studies for estimating dense deformation fields from a static MRI scan to a set of low-resolution dynamic MRI scans are presented in~\cite{makki2018high,makki2019temporal,makki2019vivo} in the context of articulated polyrigid registration of human joints. Another study to perform a high-resolution temporal reconstruction of the bladder during loading exercises is recently introduced in~\cite{ogier20193d}. Promising results have been obtained, showing the bladder in its 3D complexity during deformation due to strain conditions with an estimation of the most deformed tissue areas.

In this work, we propose to exploit the full potential of such reconstructed data for characterizing the bladder deformations in the high-resolution domain. The idea behind is to parameterize the reconstructed organ surface with a reduced number of points representing the vertices of a temporal quadrilateral mesh in order to use the generated (3D+time) smooth mesh sequence for finite element simulations. This configuration is then required for establishing a robust biomechanical model of organ dynamics~\cite{jiang2019virtual}. This clinical context is the primary motivation of this work, the results of which go beyond this framework and can be generalized to any 3D shapes. \\
A compact statistical modelization of shape change by only a small number of parameters is also required. Since the mesh vertices can be considered as a point cloud, we analyse a set of common covariance-based features incorporating the concept of "neighborhood of a point"~\cite{pauly2002efficient,dube2016segmatch,jovanvcevic20173d,liu2019lpd}, pinpointing some limitations regarding their sensitivity to noise and their dependency on point cloud density in addition to the effects of neighborhood size. Instead, we propose a different feature type based on the otpimal geodesic paths that map the moving surface points to their corresponding points in the surrounding sphere in order to evaluate the local shape variability.\\
In~\cite{yezzi2003eulerian}, a robust Eulerian PDE approach for computing the thickness of soft tissues between two non-intersecting boundaries is introduced. In this work, we extend this approach to compute geodesic distances and shortest paths and curves from an arbitrary shape to its surrounding sphere in order to derive a new geometric feature which is invariant to \textit{scaling} and \textit{rotation}. The idea is to create a virtual differentiable manifold between the shape and its surrounding sphere to overcome the computational problems related to spatial discontinuity between sparse surface points. Obtained results on synthetic shapes demonstrate that the proposed feature outperforms traditional covariance-based features.

\section{Related works} 

\section*{Point cloud tracking}

Point cloud tracking has led to many significant advances in many computer vision applications such as robotics~\cite{pomerleau2015review}, astronomy~\cite{he2017non} and medical imaging~\cite{rahim2013diffeomorphic}. A point cloud based dynamical system modulation for reactive avoidance of concave and convex obstacles during robot motion is presented in~\cite{saveriano2013point}. A method for non-cooperative spacecraft pose tracking based on point cloud feature is presented in~\cite{he2017non}.\\
Tracking a set of points can be formulated as a registration problem where the goal is to align between two point sets representing a 3D shape at two different times. The Iterative Closest Point algorithm (ICP), originally proposed by \textit{Besl} and \textit{McKay}~\cite{besl1992method} for estimating a global linear transformation to align two point sets, has been extended for non-linear registration of large 3D point sets based on a statistical expectation-maximisation (EM) algorithm~\cite{combes2020efficient}. However, robustifying this algorithm face to local minima problems occuring during the optimization of cost function remains a challenging task. The LDDMM registration framework~\cite{bone2018learning} has made it possible to estimate smooth continuous-time trajectories mapping between two point clouds without having to use point-to-point correspondence. This Riemannian-based framework can deal with large deformations, yielding geometric transformations characterized with a list of nice algebraic properties such as \textit{smoothness}, \textit{differentiability}, and \textit{invertibility}. The LDDMM introduced in~\cite{beg2005computing} has been extended to several applications: a Bayesian atlas application is proposed by \textit{Gori et al.} in~\cite{gori2017bayesian}. A geodesic shape regression with multiple geometries and sparse parameters is presented in~\cite{fishbaugh2017geodesic}. \textit{Bone et al.} have detailed a scheme for parallel transport on a high-dimensional manifold of diffeomorphisms based on the LDDMM~\cite{bone2018learning}, in the context of shape analysis. In this work, we propose to employ the LDDMM to estimate a smooth continuous curves of highly-deformable shapes represented with a set of sparse parameters (\textit{i.e.} the surface $S$ is sampled using a set of points $\{x_i\}_{i=1}^N$). 

\section*{Feature extraction from 3D point cloud data}
Feature extraction is a crucial process in every knowledge representation and classification~\cite{bohi2017fourier,mao2019interpolated}. In particular, most studies working on point clouds are based on an eigenanalysis method employing features derived from a local structure tensor, \textit{i.e. local covariance matrix} calculated from each point neighborhood. Local covariance matrices represent second-order invariant moments within the point positions~\cite{pauly2002efficient,nurunnabi2015outlier,jovanvcevic20173d}.
The local covariance matrix is defined on a k-neighbor\\-hoods of a given point $p$ according to:
\begin{equation}
C(p)= \frac{1}{k}   \sum_{i=1}^{k} (p_i - \overline{p_i})(p_i - \overline{p_i})^T
\end{equation}
where $\{p_i\}_{i \in 1 \ldots k}$ are the $k$ nearest neighbor points to the point $p$, and $\overline{p_i}$ holds the centroid of the neighbors $\{p_i\}_{i \in 1 \ldots k}$.

The obtained covariance matrix at each point is symmetric and positive semi-definite. The eigenvectors $\{e_1, e_2, e_3\}$ of the covariance matrix together with the corresponding eigenvalues $\{\lambda_1, \lambda_2, \lambda_3\}$ allow to locally estimate the surface variation. Assuming that the eigenvalues of $C(p)$ are sorted as follows: $0 \leq \lambda_1(p) \leq \lambda_2(p) \leq \lambda_3(p)$, the covariance-based features are the measures listed below:

\begin{equation}
     \begin{split}
Anisotropy &: \quad  A_\lambda = \frac{\lambda_3 - \lambda_1}{\lambda_3}; \quad Omnivariance : \quad  O_\lambda = \sqrt[3]{\lambda_1\lambda_2\lambda_2}\\
Linearity &: \quad L_\lambda = \frac{\lambda_3 - \lambda_2}{\lambda_3}; \quad \quad \quad Planarity : \quad  P_\lambda = \frac{\lambda_2 - \lambda_1}{\lambda_3} \\
Sphericity &: \quad  S_\lambda = \frac{\lambda_1}{\lambda_3}; \quad \quad \quad \quad \quad Curvature : \quad  C_\lambda = \frac{\lambda_1}{\lambda_1 + \lambda_2 + \lambda_3}
     \end{split}
\end{equation}

All or some of these geometric features have been used in many applications such as point-sampled surfaces simplification~\cite{pauly2002efficient}, 3D point cloud classification~\cite{weinmann2017geometric}, and recently used for detection and characterization of defects on airplane exterior surfaces~\cite{jovanvcevic20173d}.

The key idea of this approach is to derive these covariance features from a suitably chosen neighborhood size. However, the optimal neighborhood size cannot be predictable~\cite{zhao2019pointweb} and the proposed features may not be robust and effective enough for some complex geometries as illustrated in the example of a deformed sphere (Fig.~\ref{robust_metric}).\\

The geodesic distance is of great importance for solving many computer vision problems such as surface segmentation into regions called Voronoi cells, sampling of surface points at regular geodesic distance, and meshing a 3D surface with geodesic Delaunay triangles. In the work of \textit{Peyr\'{e} et al.}~\cite{peyre2010geodesic}, the optimal geodesic curve mapping between two surface points is obtained using Riemannian metric tensors by integrating an ordinary differential equation (ODE) modelizing the spatio-temporal evolution of surface variation using the Riemannian Eikonal equation. In this work, we propose a reverse Eulerian  approach for deriving the surface variation from optimal geodesic curves mapping  between each surface point and the surrounding sphere by integrating a pair of PDEs based on the solution of Laplace's equation, constrained by Dirichlet boundary conditions. The obtained feature is inversely proportional to the length of the geodesic curves mapping the shape to the sphere, which  maintains the bijectivity property (i.e. the uniqueness of each geodesic curve that map each surface point to its corresponding point in the sphere).\\

\section{Methods}

\subsection{Dynamic quadrilateral mesh}

\subsubsection{Quadrilateral mesh generation.}

\begin{figure*}[!h]
\centering
%\begin{minipage}{0.36\textwidth}
\subfigure{\includegraphics[scale=0.22]{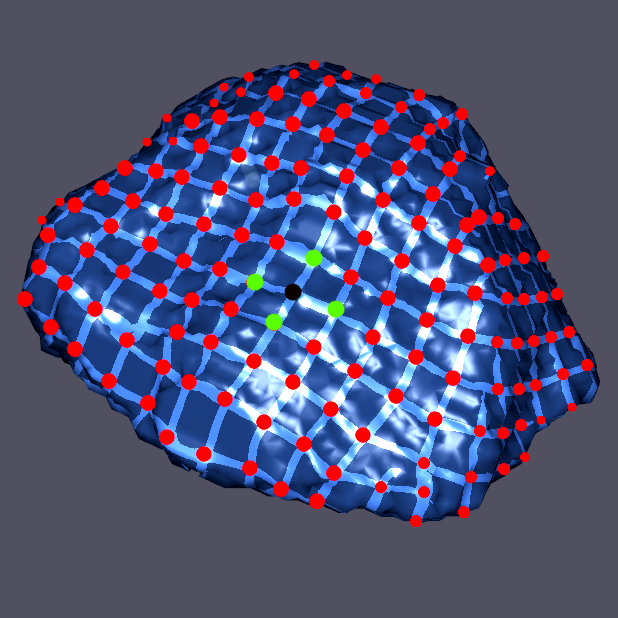}}
\hspace{1cm}
\subfigure{\includegraphics[scale=0.22]{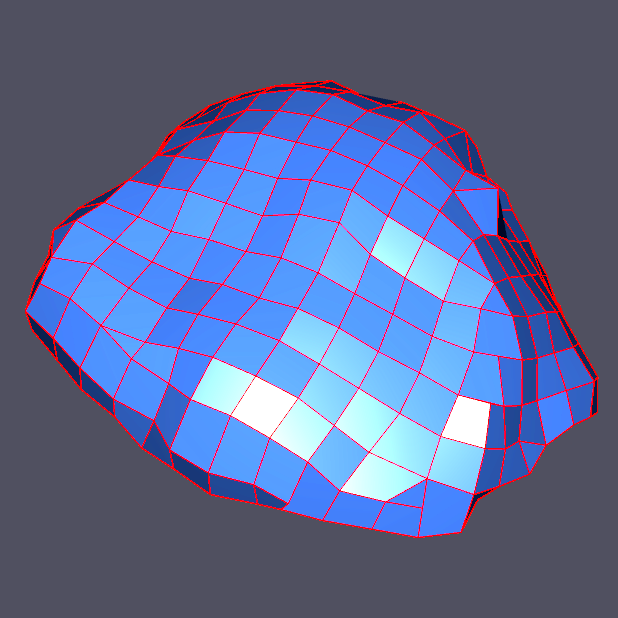}}
\caption{\label{spatial_parameterization} Quadrilateral mesh for the organ, from left to right: the position field, and the output mesh.}
\end{figure*}

To take into account the complexity of the organ shape, a topologically regular quadrilateral mesh of the organ surface is generated in the first reconstructed image in the sequence, using a robust algorithm recently presented in~\cite{jakob2015instant}. This algorithm is robust enough to establish a convex quadrilateral mesh for the organ surface since it avoids irregularity problems at the poles, as encountered in~\cite{chen2015female}, despite some singularities which can be regularized by mesh filtering in order to obtain a pure quadrilateral mesh. 
Fig.~\ref{spatial_parameterization} illustrates the quality of the obtained mesh. \\
Then, we propose to track the mesh vertices while preserving their connections (\textit{i.e.} while keeping the faces unchanged). This allows for constructing a spatio-temporal structured meshes which might be used later for deriving some biomechanical properties of the organ dynamics such as distortion, elongation and stresses using finite element methods which are required for establishing a biomechanical model for the organ dynamics. The tracking process will be detailed in the next section. 

\subsubsection{Estimation of smooth vertex trajectories}

In this section, we present a methodology to track the set of mesh vertices during the bladder deformations. This is done using the LDDMM framework that has been heuristically shown to produce natural deformation paths. A smooth and continuous-time trajectory of each vertex is estimated throughout the organ range of motion.  
The LDDMM framework can be used for manipulating dense imagery and for tracking a dynamic set of feature points over a long dynamic MRI sequence.
The bladder deformation can be interpreted and parameterized using a control-points-based LDDMM method for constructing diffeomorphisms of the ambient space $\mathbb{R}^3$. This diffeomorphic registration estimates a smooth and invertible deformation field which maps a shape to another without having to use point-to-point correspondence.

The principle of control-points-based LDDMM for estimating a diffeomorphic mapping is as follows: \\
Given a set of $N$ control points $\{q_i\}_{i \in 1,\ldots,N}$, and a set of $N$ corresponding momentum vectors of $\mathbb{R}^3$  $\{\mu_i\}_{i \in 1,\ldots,N}$, the velocity vector field which is a simple representation of the deformation vector field in the tangent space, is obtained through the use of a convolution filter:

\begin{equation}
    v : x \in  \mathbb{R}^3 \mapsto v(x)=\sum_{i=1}^{N}K(x,q_i).\mu_i
    \label{gauss_kernel}
\end{equation}
where $K(x_i,x_j) = exp(-||x_i-x_j||^2/ \sigma ^2)$ is a gaussian kernel.

The temporal evolution of the organ velocity vector field can be modeled by the following Hamilton's equations of motion:

\begin{equation}
\begin{cases} \dot{q}(t) = K(q(t),q(t)).\mu(t) \\ \dot{\mu}(t) = -\frac{1}{2} \nabla_q \{K(q(t),q(t)), \mu(t)^{\top}\mu(t)\} \end{cases}
\end{equation}

Solving this pair of PDEs using a second-order Runge-Kutta scheme gives a smooth temporal velocity vector field derived from $q(t)$ and $\mu(t)$  :

\begin{equation}
    v : x \in  \mathbb{R}^3 \times t \in [0,1] \mapsto  v(x,t)=\sum_{i=1}^{N}K(x,q_i(t)).\mu_i(t)
\end{equation}

The temporal evolution of each tracked point $x \in \mathbb{R}^3$  is governed the following ODE:

\begin{equation}
   \dot{x}(t) = v(x(t),t) 
\end{equation}

with the initial condition $x(0)=x$.\\

Finally, the solution of this ODE yields a flow of diffeomorphisms $\Phi_{q,\mu}(x,t) :  \mathbb{R}^3 \mapsto \mathbb{R}^3$, such that $\Phi_{q,\mu}(x,1)$ is the desired deformation vector field that maps between source and target.

\begin{figure*}[h!]
\centering
%\begin{minipage}{0.36\textwidth}
\subfigure{\includegraphics[scale=0.124]{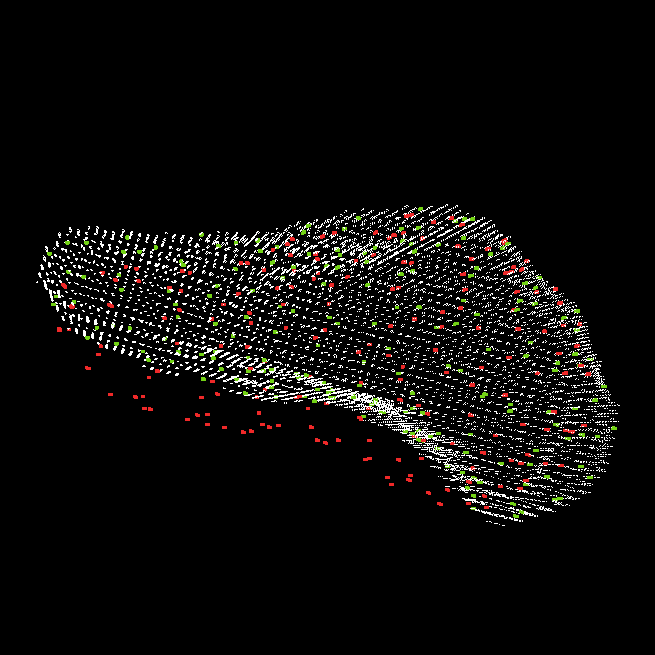}}
%\subfigure{\includegraphics[scale=0.26]{eccv2020kit/mean_curvature.png}}
\hspace{0.01cm}
\subfigure{\includegraphics[scale=0.124]{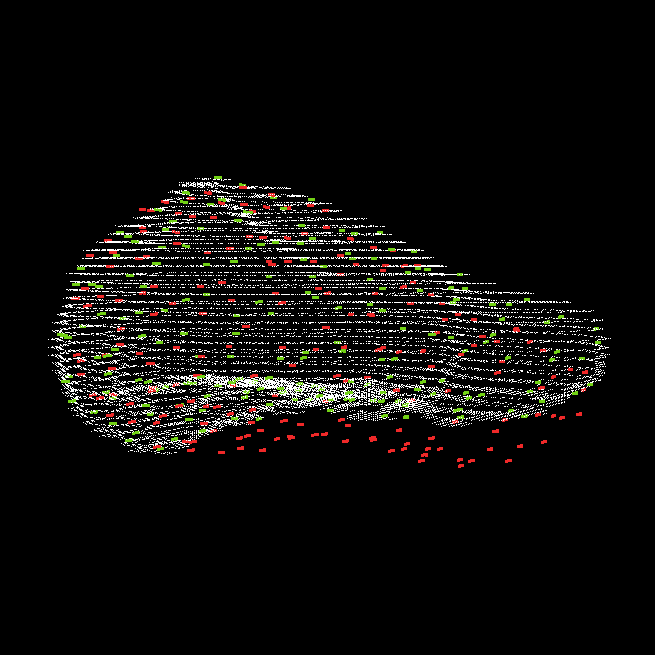}}
\hspace{0.01cm}
\subfigure{\includegraphics[scale=0.124]{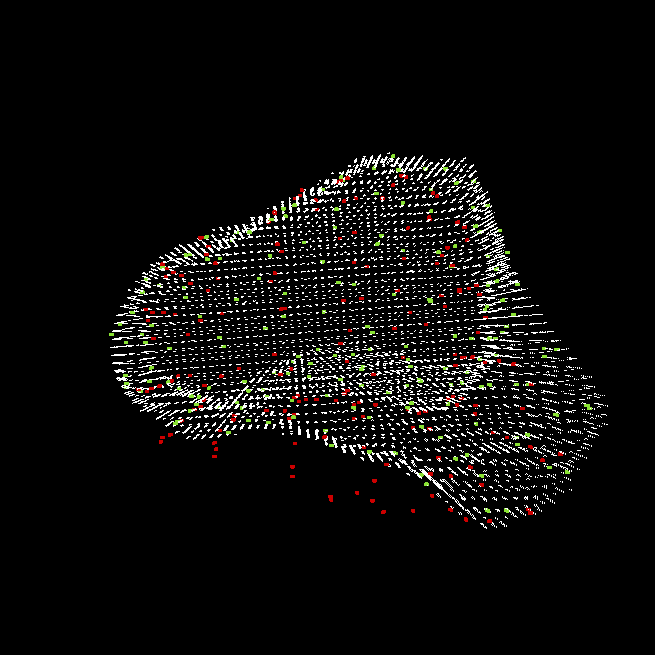}}
\hspace{0.01cm}
\subfigure{\includegraphics[scale=0.124]{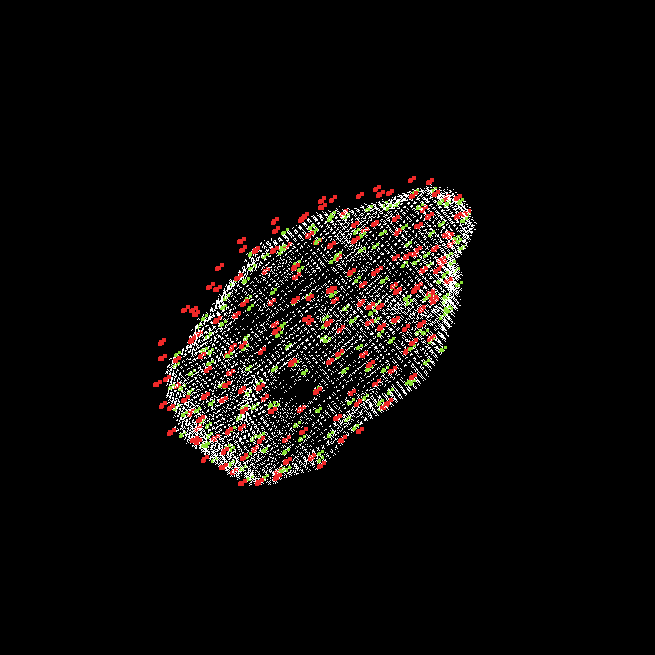}}
    \caption{Pointset tracking: : $\mathcal{M}_0$ in red, $\mathcal{M}_t$ in green, and $\mathcal{C}_t$ in white.}
    \label{fig:tracking_algorithm}
\end{figure*}

% \subsection*{Application}
The overall algorithm for point tracking is described in Algorithm~\ref{algo:motion}, with the following notations:  
$L$ is the length of the dynamic sequence; $\mathcal{M}_t$ gives the locations of the point set being tracked at time $t$; $\mathcal{C}_t$ is the 3D contour point cloud at time $t$; where $\mathcal{M}_t$ is a proper subset of $\mathcal{C}_t$. Note that the registration problem is solved by iteratively minimizing the following loss function:
\begin{equation}
    f(q,\mu)= d(\mathcal{C}_{t+1} , \Phi_{q,\mu}(\mathcal{M}_t)) + R(q,\mu)
    \label{loss_function}
\end{equation}
where the first term measures data-attachment while the second regularization term represents the norm of the deformation.
Some results of our tracking algorithm are illustrated in Fig.~\ref{fig:tracking_algorithm}.

\begin{algorithm}
\caption{Tracking of mesh vertices}\label{algo:motion}
%\begin{algorithmic}
\textbf{Input:} Mesh vertices $\mathcal{M}_0$. 

\textbf{Motion estimation:} Estimate forward successive point trajectories using the LDDMM $\{ \mathcal{M}_{t+1}\}_{t~=0,\ldots,L-1}$ such that $\mathcal{M}_t \subset \mathcal{C}_t $, for $t~=0,\ldots,L-1$, by aligning $\mathcal{M}_{t}$ and $\mathcal{C}_{t+1}$.
%\end{algorithmic}
\end{algorithm}

\subsection{Feature analysis: novel geometric feature type}

In this section, we propose a robust global feature to detect the surface variation in terms of concavity and convexity with smooth transitions in-between. For a given 3D shape $S$, the \textit{proposed} feature can be numerically approximated using an Eulerian PDE approach, involving the following steps:

\subsection*{Definition of Dirichlet boundary conditions:} To initialize the algorithm, the binary mask of the shape is eroded with a cross-shaped structuring element which is best suited for fine structures. The choice of the structuring element is of great importance for preserving, as much as possible, the topology of any arbitrary shape. This gives the eroded binary mask that we note $S_{e}$. 
Then, we use a PCA analysis of $S$ in order to determine the radius $R$ of the surrounding sphere $S_s$ centered at the shape centroid. The radius $R$ is proportional to the length of the principal axis of inertia $l$ ($R= 0.8l$). At this level, all the shape contour points will be located between two non-intersecting boundaries: $S_{in}=S_e$, and $S_{out}=\bar{S_s}$.\\
The next step consists of computing the length of the shortest geodesic paths from $S_{in}$ to $S_{out}$ in a bijective fashion. The full algorithm is described below:

\subsection*{Solving the Laplace equation, subject to Dirichlet boundary conditions:} Determine a twice-differentiable function $h :  \mathbb{R}^3  \rightarrow  \mathbb{R}$ which satisfies $\Delta h = div(grad(h)) = 0 $ inside the region $\Omega = \overline{S_{in} \cup S_{out}}$, subject to the Dirichlet boundary conditions $\partial S_{in}=0$ and $\partial S_{out}=10^4$. To approximate the numerical solution of Laplace equation, we use the Jacobi iterative relaxation method:
    \begin{equation}
    \begin{matrix}
    h_{t+1}(i,j,k)=\frac{1}{2(d_j^2d_k^2+d_j^2d_k^2+d_i^2d_j^2)}(d_j^2d_k^2[h_{t}(i+d_i,j,k)+h_{t}(i-d_i,j,k)]\\
    +d_j^2d_k^2[h_{t}(i,j+d_j,k)+h_{t}(i,j-d_j,k)]+d_i^2d_j^2[h_{t}(i,j,k+d_k)+h_{t}(i,j,k-d_k)])\\
    \end{matrix}
    \end{equation}\\
    
    where $t$ is the iteration index. In this work, we process reconstructed data with isotropic voxel spacing of $1 \times 1\times 1 mm$ (\textit{i.e.} $d_i = d_j = d_k = 1$).\\
    
\subsection*{Computing the normal vectors to the tangent planes of the harmonic layers:} Determine elementary normal paths from $S_{in}$ to $S_{out}$, \textit{i.e.} from the normalized gradient vector field of the harmonic interpolant $h$, $\overrightarrow{T}= \frac{\nabla h}{\left \| \nabla h \right \|}$. 
\subsection*{An Eulerian PDE computational scheme:} We estimate the optimal geodesic paths between the two boundaries, by solving a couple of PDEs: $\nabla L_0.\overrightarrow{T} = -\nabla L_1.\overrightarrow{T} = 1$. An appropriate symmetric Gauss-Seidel relaxation method is used for numerical integration for this pair of PDEs:\\
    
\begin{equation}\label{L0}
    L_0^{t+1}[i,j,k] = \frac{1+\left | T_i \right |L_0^t[i\mp1,j,k]+\left | T_j \right |L_0^t[i,j\mp1,k]+\left | T_k \right |L_0^t[i,j,k\mp1]}{\left | T_i \right |+\left | T_j \right |+\left | T_k \right |} 
\end{equation}
\begin{equation}\label{L1}
    L_1^{t+1}[i,j,k] = \frac{1+\left | T_i \right |L_1^t[i\pm1,j,k]+\left | T_j \right |L_1^t[i,j\pm1,k]+\left | T_k \right |L_1^t[i,j,k\pm1]}{\left | T_i \right |+\left | T_j \right |+\left | T_k \right |} 
\end{equation}

where: $\left\{
\begin{array}{l}
  m \pm 1 = m + sgn(T_m) \quad ; \quad  m \mp 1 = m - sgn(T_m)  \\
  m \in{\{i,j,k\}}  \\
\end{array} \right.\\$

    with: $ sgn(.) $ is the sign function, $L_0[i,j,k]$ is the length of the optimal geodesic path from the point $x=(i,j,k)$ to $S_{in}$, while $L_1[i,j,k]$ is the length of the optimal geodesic path from $x$ to $S_{out}$. 
The sum of these two lengths $G(x) = L_0(x)+L_1(x)$, defined as \textit{thickness} in~\cite{yezzi2003eulerian}, represents in fact the length of the optimal geodesic path from $S_{in}$ to $S_{out}$ that passes through $x$.

\subsection*{Proposed feature to characterizing the surface variation:} 
\begin{figure}[!h]
    \centering
    \includegraphics[width=0.9\linewidth]{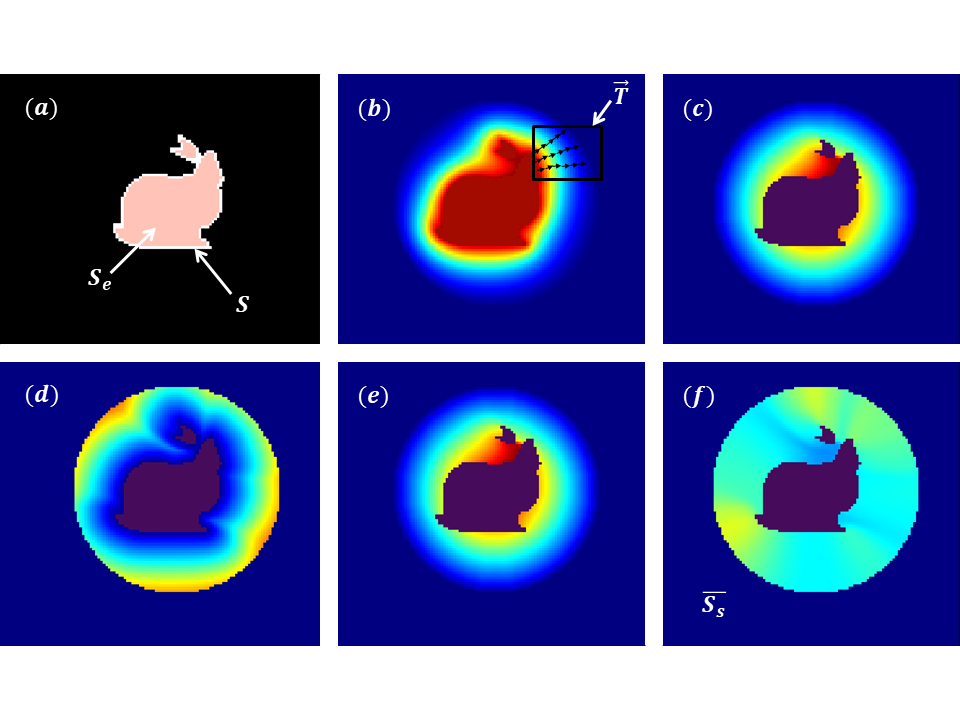}
    \vspace{-0.6cm}
    \caption{Proposed method: (a) binary mask and its eroded; (b) solution of the Laplace equation $h$; (c) length of the geodesic path from $x$ to the inner boundary $L_0$; (d) length of the geodesic path from $x$ to the surrounding sphere $L_1$; (e) length of the resulting geodesic path $G$; and (f) feature values $\tilde{f}$ between the two boundaries.}
    \label{fig:proposed_method}
\end{figure}

Finally, we define a flexible feature by the the following application $\tilde{f} :  \mathbb{R}^3  \rightarrow  \mathbb{R}$:
    \begin{equation}
        \tilde{f}(x) = \frac{R}{G(x)}
    \end{equation}
The function $\tilde{f}$ has the same potential to the Reimannian curvature for characterizing surface variation and for smoothly delineating between concave, convex and flat regions as illustrated in Fig~\ref{ellipsoid_curv} for both torus and ellipsoid. The function $\tilde{f}$ gives a bijective way of characterizing surface variation since the geodesic paths may never intersect.This suggests that the curvature of a given 3D shape at a point $x$ is inversely proportional to the length of the geodesic curve from the shape to its surrounding sphere, that passes through $x$. Relatively, the largest feature values correspond to the most convex areas in the surface while the smallest values correspond to the most concave areas. The proposed feature values exhibit smooth transitions between concave and convex regions of the surface. Furthermore, this feature is invariant to scaling since the sphere radius is proportional to the length of the principal axis of inertia of the shape and also invariant to rotation thanks to spherical symmetry. An illustration of all the previous steps is presented in Fig.~\ref{fig:proposed_method} for the example of the Stanford Bunny.

\begin{figure}
    \centering
    \includegraphics[width=1.0\linewidth]{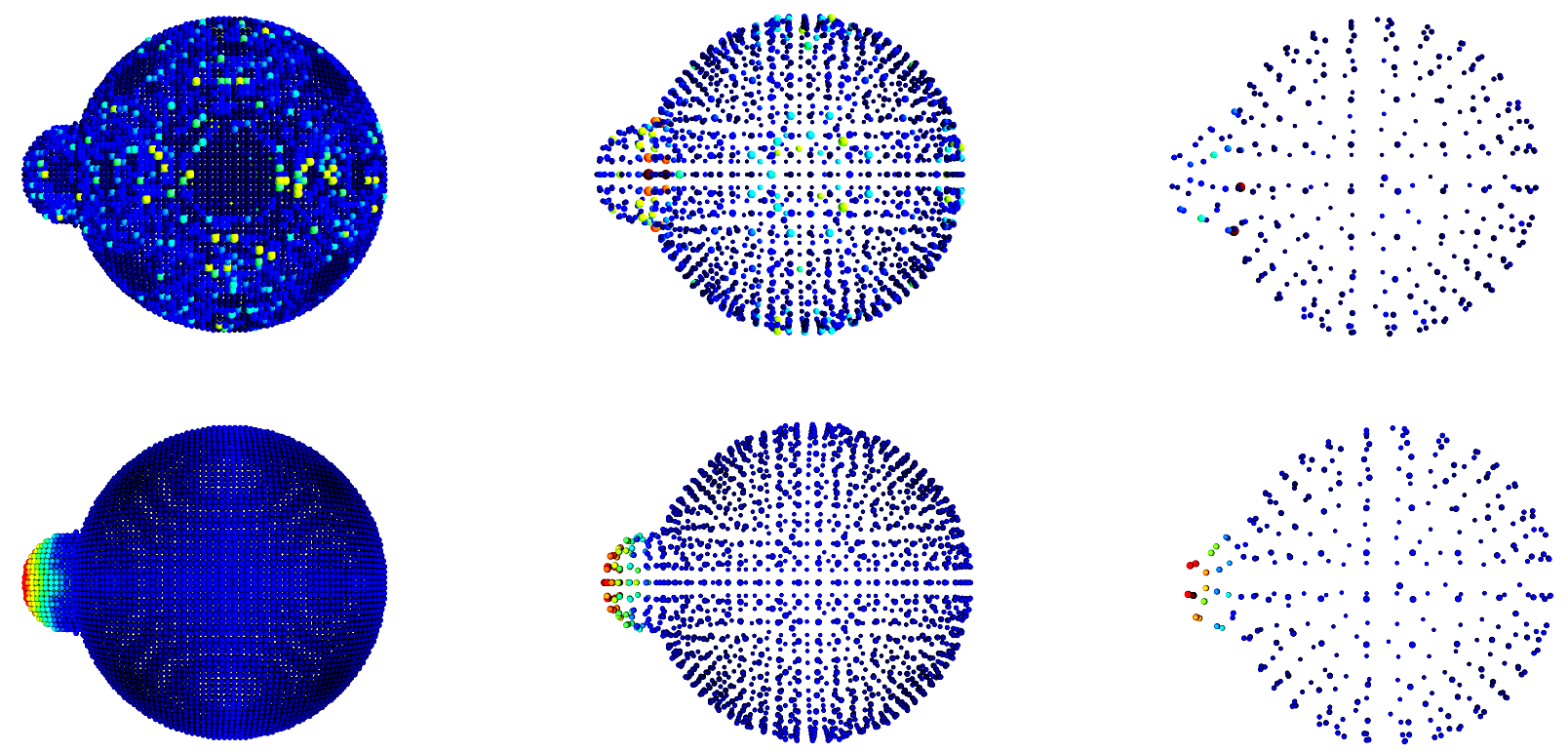}
    \caption{\label{robust_metric} Local characterization of 3D surfaces. First row: the covariance-based shape curvature with 4-nearest neighbors. Second row: our proposed feature based on the geodesic distances. The columns show the robustness of the proposed metric face to the point cloud density: first column: voxel size equal to 1 (\textit{i.e.}, we keep all the contour points), second column: with a voxel size of 3, and with a voxel size of 7 in the third column.}
\end{figure}

\section{Results}

\subsection{Data set}
Pelvis areas of six healthy participants were imaged with a 1.5T MRI scanner (MAGNETOM Avanto, Siemens AG, Healthcare Sector, Erlangen, Germany) using a spine/phased array coil combination. T$_{1}$/T$_{2}$W bSSFP images (TR: 125 ms, echo time: 1.25 ms, flip angle: 52$^{\circ}$, field of view: 299 x 350 mm$^2$, pixel size: 1.36 x 1.36 mm$^2$, slice thickness: 6 mm, multi-planar configuration) were recorded during a 1:20-minutes forced breathing exercise. During this exercise, the subject alternately inspired and expired at maximum capacity. Subjects were also instructed to increase the pelvic pressure to the maximum inspiration and contrary to contract the pelvic floor during the expiration. These actions increased the intra-abdominal pressure, causing deformities of the pelvic organs. The study was approved by the local human research committee and was conducted in conformity with the declaration of Helsinki. Since no extrageneous liquids was injected into pelvic cavities in this study, only the segmentation of the bladder was performed and the analysis focused exclusively on this organ. For each subject, the three-dimensional dynamic sequences acquired in multi-planar configurations allowed the reconstruction of nearly 400 bladder volumes generated at a rate of 8 volumes per second.\\

\subsection{Validating the point-tracking process}

The parameters of our registration algorithm are set as follows: the standard deviation of the Gaussian kernel defined in~\eqref{gauss_kernel} is set to $\sigma = 10^{-4}$ in order to obtain a deformation field with a very thin level of precision. The kernel-width parameter for controlling the granularity of the deformation is set to $8$. $15$ intermediate states describing the temporal evolution of the tracked points are estimated in order to obtain a smooth continuous-time trajectory of the organ between successive time frames since the registration process is much simpler to converge for transformations very close to the identity. The loss function defined in~\eqref{loss_function} is minimized using the gradient descent optimization method.\\
All experiments are performed on an Intel$^{\mbox{\scriptsize{\textregistered}}}$  Xeon$^{\mbox{\scriptsize{\textregistered}}}$ Processor Silver 4214 CPU \verb+@+ 2.20GHz, with a physical memory of 93GB. For the first subject for example, Algorithm~\ref{algo:motion} takes $6$ sec to align a set of $342$ tracked point set $\mathcal{M}_t$, with the target set $\mathcal{C}_t$, composed of $5686$ points for which only a set of $210$ contour points have been used for optimizing the shape matching.

To validate the tracking process, we propose to compute the following error:

\begin{equation}
  E = \frac{1}{N} \sum_{p=1}^{N} dist(x_p,\mathcal{C}_L)
\end{equation}

where: $N$ is the total number of tracked points; $x_p \in \mathcal{M}_L$; and $dist(x_p,\mathcal{C}_L)$ is the Euclidian $\ell^2$ distance between $x_p$ and the closest point $x_l$ in the last reconstructed contour $\mathcal{C}_L$. The results for method evaluation are reported in Table~\ref{tab1}.

\begin{center}
\begin{table}[!h]
 \caption{Tracking/interpolation error.\label{tab1}}
\begin{tabular*}{\textwidth}{c @{\extracolsep{\fill}} ccccccc}
  \hline
  Subject & S1 & S2 & S3 & S4 & S5 & S6\\
  \hline
 Error & 0.58 & 0.56 & 0.69 & 0.58 & 0.66 & 0.62\\
 \hline
\end{tabular*}
\end{table}
\end{center}

\begin{figure}[!h]
    \centering
    \includegraphics[width=1.0\linewidth]{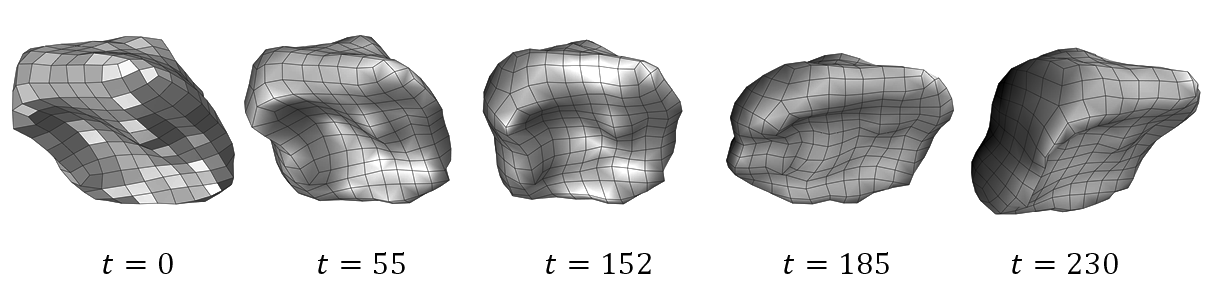}
    \vspace{-0.7cm}
    \caption{Construction of smooth 4D quadrilateral mesh during forced respiratory motion.}
    \label{fig:4Dmesh}
 \end{figure}

Table~\ref{tab1} shows a mean propagated error inferior to $1 mm$ which reflects the accuracy of the results regarding isotropic voxel size of $1\times1\times1 mm$. Fig.~\ref{fig:4Dmesh} shows the quality of our 3D+t quadrilateral mesh reconstruction based on smoothly tracking vertices using the LDDMM while keeping the mesh faces unchanged.

\subsection{Feature application to synthetic data}

\begin{figure}[!t]
    \centering
    \includegraphics[width=0.9\linewidth]{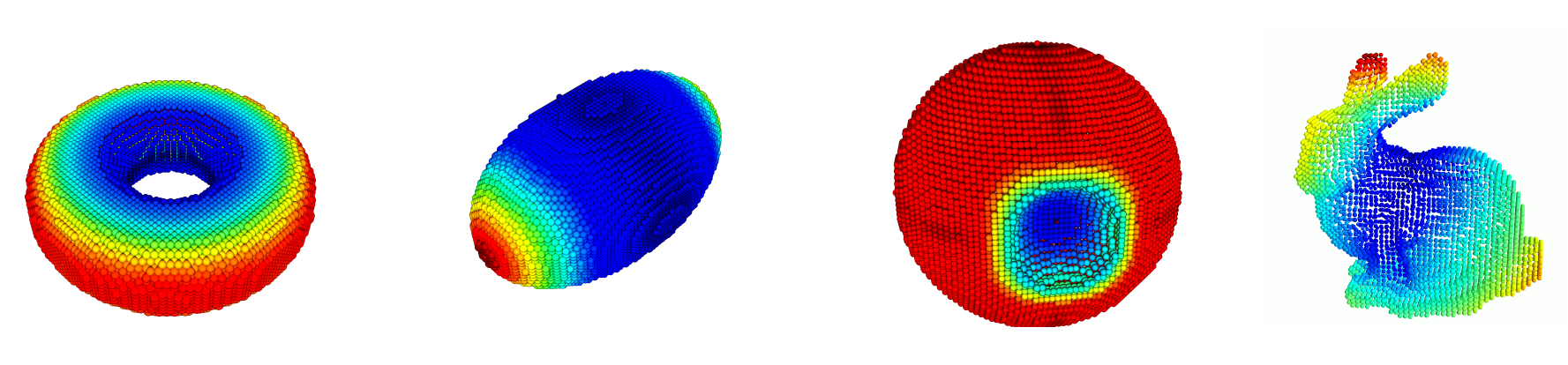}
    \vspace{-0.3cm}
    \caption{\label{ellipsoid_curv} Application to simulated geometric forms, from left to right: for a torus, for an ellipsoid, for a sphere with closed gap , and for the Stanford Bunny.}
\end{figure}

The performances of our 3D geometric descriptor are evaluated on some known synthetic surfaces, (see Fig.~\ref{ellipsoid_curv}). 
Since all the covariance-based features, detailed in the related works section, describe the local surface variation, we compared our 3D geometric descriptor with one of these features because the main idea is rather to show the robustness to the change in point cloud density since the covariance-based features are mainly based on the optimal definition of point neighborhood. The covariance-based curvature is computed using 4-nearest neighbors. Each point neighborhood is determined using a ball tree algorithm. An example of a point neighborhood is illustrated in Fig~\ref{spatial_parameterization} (the black point and its 4-nearest neighbors in green).   
Fig.~\ref{robust_metric} shows that the covariance-based features become more sensitive to noise with an increase of the number of surface points.

\subsection{Characterization of the bladder deformations}

After validating the tracking process of bladder points and presenting the data set used in our study, we analyse here some experimental results to show an example of application.\\

\begin{figure}[!t]
    \centering
    \includegraphics[width=0.8\linewidth]{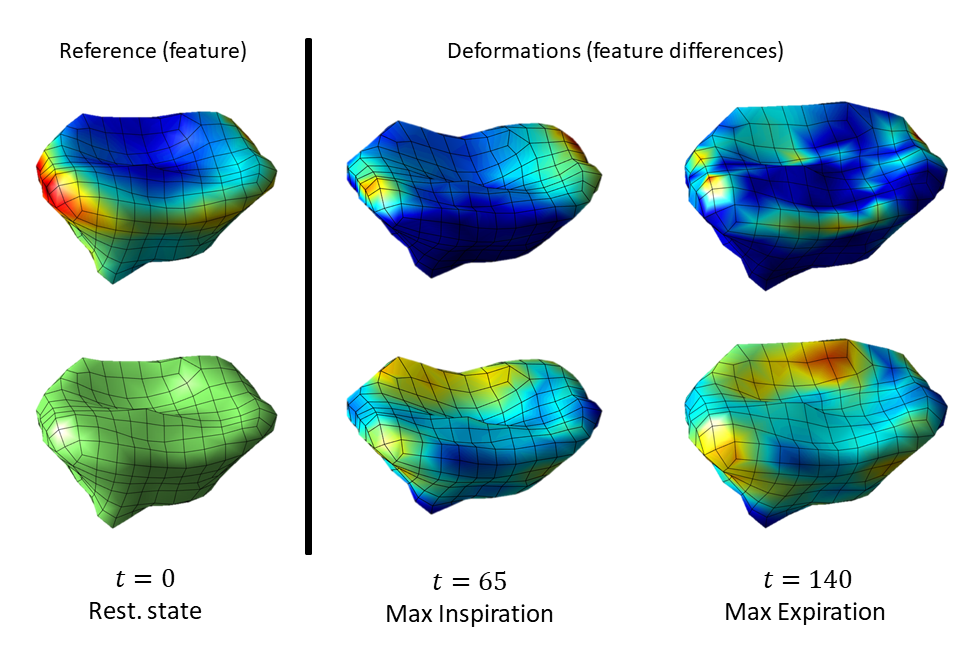}
    \vspace{-0.3cm}
    \caption{\label{curvature_dyn} Characterization of the organ shape variability at mid and extreme range of motion: The first row shows the texture mapping of our geometric feature on bladder meshes, while the second row shows the mesh elongations. Colormap goes from blue (-) to red (+).}
\end{figure}

Fig.~\ref{curvature_dyn} illustrates the characterization of the organ shape variability at mid and extreme range of motion: resting state $t=0$, maximum of inspiration $t=65$ causing a huge organ contraction, and maximum of expiration $t=140$ inducing an important bladder swelling.   
The mapping of our feature as a texture on the reconstructed quadrilateral mesh surface is presented in first row, first column of the figure. Second and third columns show feature differences with respect to the reference resting state, highlighting the regions affected by the deformations.
The second row of the figure shows the mapping of mesh elongations on the surface w.r.t. the reference resting state. The elongation also called the Green-Lagrange deformation is a commonly used feature for characterizing local deformations and it measures the level of the spacing in the neighborhood of a point.

\begin{figure}[!t]
\centering
\subfigure{\includegraphics[scale=0.26]{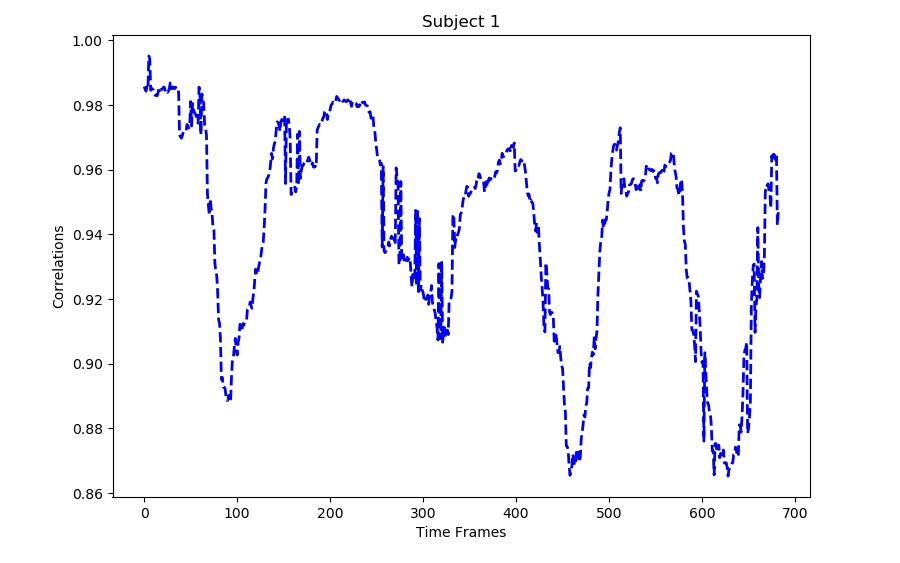}}
\subfigure{\includegraphics[scale=0.26]{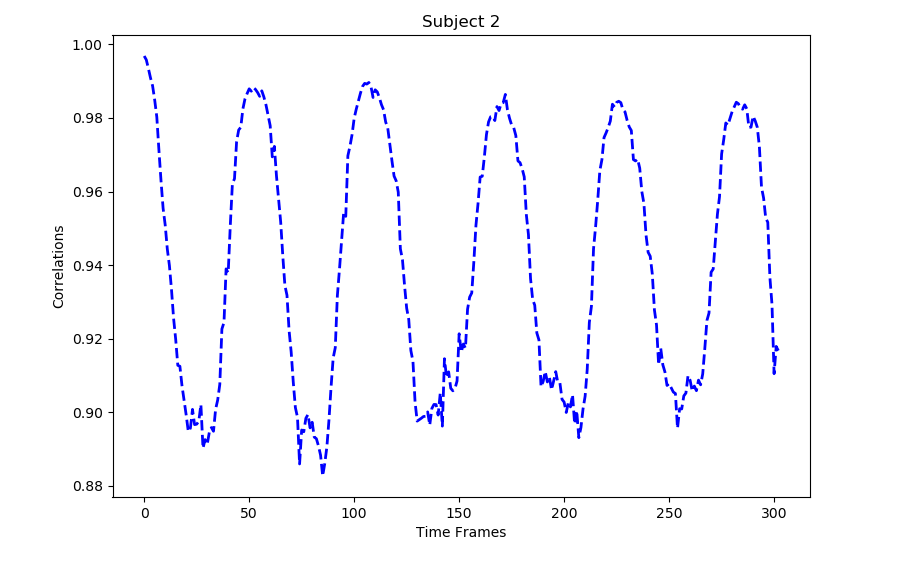}}
%\subfigure{\includegraphics[scale=0.26]{eccv2020kit/AF_graph.png}}
\vspace{-0.3cm}
\caption{\label{correlations} Correlations between each temporal feature map and the initial one for two subjects.}% AF and AR}% and CM}
\end{figure}

In order to characterize the bladder deformations during loading exercises, we compute our invariant geometric descriptor on the surface subjects, then using Pearson correlation metric, we compare each feature map at a time frame $t$ with that of the initial one ($t=0$).
Results presented in Fig.~\ref{correlations} illustrate variations in the correlation between each temporal feature map w.r.t the feature map in the first time frame for two subjects S1 and S2. A decrease of the correlation value indicates organ deformation with respect to the reference (initial shape), this can be interpreted by the fact that forced inspiration involves an action of the diaphragm and abdominal muscles that induces deformations of the internal organs. Otherwise, when the correlation values increase, the patient releases the pressure and consequently the bladder relaxes and returns to its initial shape ($t=0$).
Taking for example the case of the second subject, for time frames $t\in[0,30]$, we note a correlation drop down from 1 to 0.9 characterized by an inspiration mode. Beyond $t=$30 to 50, the correlation values increase reflecting the fact that the subject is in expiration mode. So on, the correlations between the shapes of the bladder vary according to the breathing mode applied by the patient. We note here the noise present in our graphs, which is simply caused by errors of 3D reconstruction of the sequences from dynamic MRI and also due to the instability of the respiratory motion for each subject.\\

\begin{figure}[!h]
    \centering
    \subfigure{\includegraphics[scale=0.25]{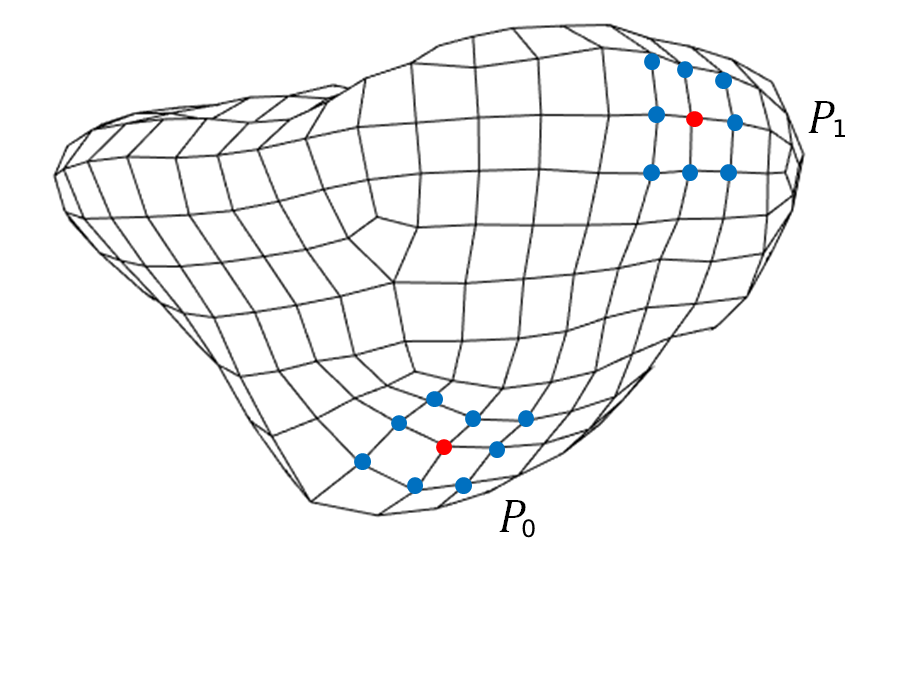}}
    \subfigure{\includegraphics[scale=0.33]{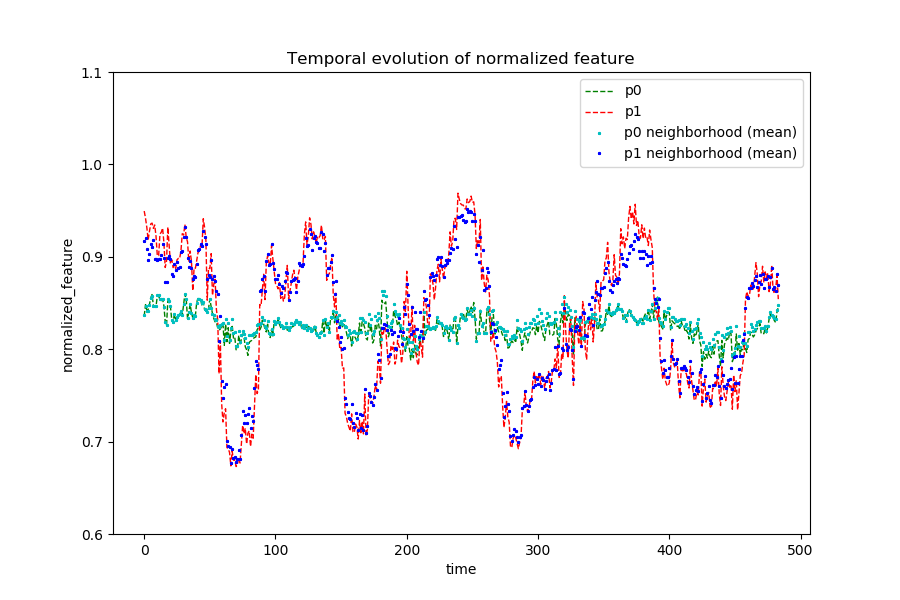}}
    \vspace{-0.5cm}
    \caption{Normalized feature evolution around two significant points.}
    \label{fig:point_feature}
 \end{figure}

Fig.~\ref{fig:point_feature} illustrates the temporal evolution of feature values for two surface regions (subject S1): a first point $p_0$ is selected from a statistically non-highly deformable surface area (the bladder neck supported by the pelvic floor muscle). The average feature values in the 8-neighborhood of $p_0$ are then calculated to evaluate the local behavior of the surface variation around that point. A second point $p_1$ is selected from a highly deformable lateral region (the apex) so that the corresponding temporal feature curve fluctuated over time during forced respiratory motions. Thus, our geometric descriptor allows to separate between surface regions undergoing different deformation rates with respect to the neutral position (resting state). 

\section{Conclusion}
In this paper, we have developed a framework for characterizing the temporal evolution of three dimensional shapes and we have showed  an example of application with data acquired from the observations of an organ \textit{in vivo} using dynamic MRI. A set of contour points are first tracked using non-linear diffeomorphic registration. Second, a new geometric feature is proposed to deal with large deformations by creating a virtual differentiable manifold between the organ surface and its surrounding sphere. Results demonstrate the robustness of our descriptor no matter how sparse the point cloud may be. Future works will include the calculation of different biomechanical parameters from the obtained dynamical quadrilateral meshes such as distortions, strains and stresses. Moreover, determining the best sampling rate of surface points will be investigated.

\section*{Acknowledgement}
This research was supported in part by the AMIDEX - Institut Carnot STAR under the Pelvis3D grant.

\bibliographystyle{splncs04}
\bibliography{egbib}
\end{document}